\documentclass[conference]{IEEEtran}
\ifCLASSINFOpdf
\else
\fi
\usepackage[acronym]{glossaries}
\usepackage{graphicx}
\usepackage{multirow}
\usepackage{diagbox}
\usepackage{longtable}
\usepackage{array}
\usepackage{rotating}
\usepackage{eqparbox}
\usepackage{makecell, caption, booktabs}

\newcommand{\ycbcr}{$\text{Y'C}_\text{b}\text{C}_\text{r}$}
\begin{document}
\title{Time-Aware Face Anti-Spoofing with Rotation Invariant Local~Binary~Patterns and Deep Learning}

\author{
\IEEEauthorblockN{Moritz Finke}
\IEEEauthorblockA{University of Würzburg\\
moritz.finke@uni-wuerzburg.de}
\and
\IEEEauthorblockN{Alexandra Dmitrienko}
\IEEEauthorblockA{University of Würzburg\\
alexandra.dmitrienko@uni-wuerzburg.de}}


%Telephone: (800) 555--1212, Fax: (888) 555--1212}


%\IEEEoverridecommandlockouts
%\makeatletter\def\@IEEEpubidpullup{6.5\baselineskip}\makeatother
%\IEEEpubid{\parbox{\columnwidth}{
%}
%\hspace{\columnsep}\makebox[\columnwidth]{}}

\maketitle

\begin{abstract}
%\boldmath

Facial recognition systems have become an integral part of the modern world.
    These methods accomplish the task of human identification in an automatic,
    fast, and non-interfering way. Past research has uncovered high
    vulnerability to simple imitation attacks that could lead to erroneous
    identification and subsequent authentication of attackers. Similar to face
    recognition, imitation attacks can also be detected with Machine
    Learning. Attack detection systems use a variety of facial features and
    advanced machine learning models for uncovering the presence of attacks. In
    this work, we assess existing work on liveness detection and propose a
    novel approach that promises high classification accuracy by combining
    previously unused features with time-aware deep learning strategies.

\end{abstract}


% \fi
%%\IEEEpeerreviewmaketitle

\newacronym{rppg}{rPPG}{remote photoplesthymograpy}
\newacronym{rnn}{RNN}{Recurrent Neural Network}
\newacronym{cnn}{CNN}{Convolutional Neural Network}
\newacronym{svm}{SVM}{Support Vector Machine}
\newacronym{gan}{GAN}{Generative Adversarial Networks}
\newacronym{fld}{FLD}{Face Lifeness Detection}
\newacronym{pad}{PAD}{presentation attack detection}
\newacronym{ml}{ML}{Machine Learning}
\newacronym{kyc}{KYC}{Know Your Customer}
\newacronym{finra}{FINRA}{Financial Industry Regulatory Authority}
\newacronym{ctlstm}{C\&T-LSTM}{Color and Texture LSTM}
\newacronym{lstm}{LSTM}{Long Short-Term Memory}
\newacronym{lbp}{LBP}{Local binary patterns}

\section{Introduction}

The relationship of businesses and customers is continuously moving to the
digital space. For critical operations such as online banking, businesses must
follow \gls{kyc} guidelines and therefore validate customer
identities~\cite{finra2090}. In the digital space, this can be achieved with
face recognition systems. These systems are already widely deployed and
accepted. For adults in Germany and the United States, face recognition is
among the five most used authentication methods~\cite{statista2018de,
statista2018us}. Despite its prevalence, however, the main concern against
biometric authentication is its susceptibility to impersonation
attacks~\cite{statista2018de, statista2018us}. Indeed, face recognition systems
must incorporate \gls{pad}~\cite{iso30107-1} techniques in order to effectively
distinguish between real humans and impersonation attempts such as the
presentation of face printouts. Systems that do not consider such attack
vectors have been shown to be highly susceptible to the described
attacks~\cite{hadid2014face}.

There exists a multitude of features such as specialized camera systems for depth
measurement~\cite{grinchuk20213d} that \glspl{pad} can use to detect imitation
attempts. In many use cases such as
In-App authentication, however, \glspl{pad} are
restricted to the data of standard sensors such as phone cameras that operate within the
optical spectrum. In addition to the limited set of sensors, it also holds
that the authentication process should not obstruct the user experience and
must therefore finish within a short time span. Yet, previous work has shown
that the processing of multiple video frames is beneficial for the
classification accuracy of \gls{pad} systems. In the case of online
verification, another limiting factor is found in the amount of data that must
be submitted. To summarize, existing models are faced with a trade-off between accuracy,
authentication time, and resource requirements.

With this work we propose a \gls{pad} system termed \gls{ctlstm} that enables
quick authentication while simultaneously incorporating the benefits of
multi-frame classification and requiring only minimal bandwidth. In order to
comply with user experience, the system uses only 16 consecutive frames (0.25
to 0.5 seconds) for classification. The data that is obtained from these frames
is further minimized such that online submission does not perceptibly delay the
authentication process. The system applies a previously unused variant of the
established color texture extraction operator termed \gls{lbp} in combination
with a \gls{lstm} based deep learning model. The proposed system is evaluated
and thoroughly compared with existing work and public datasets.

Our contributions are
summarized as follows: we

\begin{itemize}
\item introduce \gls{ctlstm}, a system that provides strong attack resilience
    while being non-obstructive for authentication processes and being highly
    compatible with end-devices,
\item describe the rotation invariant uniform
pattern \gls{lbp}~\cite{ojala2002multiresolution} and its significance as a
        feature for \gls{pad},
\item show that \gls{lstm} models, that are otherwise underrepresented in
\gls{pad} systems, are well-suited for multi-frame classification, and
\item provide extensive evaluation of the proposed \gls{ctlstm} and its various
parameters along with a comparison to existing \gls{pad} systems.
\end{itemize}

In the remainder of this work we first provide an overview of related work with
Section~\ref{sec:related-work} and of relevant background information with
Section~\ref{sec:background}. The \gls{ctlstm} design is described in
Section~\ref{sec:model-design} and evaluated in Section~\ref{sec:evaluation}.
Finally, the work is concluded in Section~\ref{sec:conclusion}.

\section{Related Work}
\label{sec:related-work}

Liveness Detection has been approached with various features and
(\gls{ml}-based) algorithms. Recent work focusses on advanced deep learning
approaches such as Deep Reinforcement~\cite{cai2020drl} and
\glspl{gan}~\cite{nguyen2020presentation} as well as Domain
Adaptation~\cite{wang2021self} for unseen attacks and zero-shot
learning~\cite{khairnar2023face}.

As shown in~\cite{khairnar2023face}, earlier work uses classic deep learning
and simple \gls{ml} models such as \glspl{svm}~\cite{gundougar2021presentation,
boulkenafet2015face} and \glspl{cnn}~\cite{yang2019face}. In these cases, the
models are typically trained with (handcrafted) features. A model that uses a
rich set of features is proposed by 
G{\"u}ndo{\u{g}}ar and Erdem~\cite{gundougar2021presentation}. The authors propose a model
that fuses information about eye blinking, eye gaze, and head pose with
textural data and advanced data from \gls{rppg}
algorithms~\cite{wang-tbe-2015}. With \gls{rppg} the temporal difference of
specular and diffuse skin reflection is measured. This measurement can be used
to observe a person's blood circulation~\cite{wang2016algorithmic}. Motivated
by the idea that \gls{rppg} can be used for effectively distinguishing between
\textit{live} persons and, e.g., static printouts, the feature is also being
used for liveness detection by Liu~et~al.~\cite{liu2018learning} who use it as
supervision signal for the training process of a neural network. It must be
noted, however, that the quality of this feature relies on a large number of
consecutive frames ($>60$) and therefore can be impractical for use cases with
fast authentication and limited resources.

The idea of extracting
micro-movements such as the heart rate can also be found with approaches that
use a similar feature that is termed eulerian video
magnification~\cite{Wu12Eulerian}. This method applies
spatial decomposition using laplacian pyramids. By subsequently applying temporal filtering,
selected motion frequencies are magnified that are otherwise invisible. Besides
the heart rate, many other motion types such as breathing can
be magnified~\cite{Wu12Eulerian}. In the field of liveness detection,
Tu~et~al.~\cite{tu2019enhance} use this feature in combination with a CNN-LSTM
network whereas Zhao~et~al.~\cite{zhao2020face} pair it with the scale-invariant
feature transform (SIFT) algorithm. While it is shown that eulerian video
magnification is a capable feature for liveness detection, it must be noted
that its computationally heavy operations may make it impractical for
use cases that require lightweight or fast authentication.

Finally,
Bharadwaj~et~al.~\cite{bharadwaj2013computationally} combine eulerian video
magnification with \gls{lbp} and use a \gls{svm} for classification. The
\gls{lbp} operator (first described in~\cite{he1990texture}) is another popular feature for liveness
detection~\cite{khairnar2023face} that is also in the focus of this work. 
With \gls{lbp}, the \textit{color texture} of images can be described
efficiently. There exist multiple variants of \gls{lbp}
operators with individual characteristics such as rotation
invariance~\cite{ojala2002multiresolution}. For liveness detection, the
\gls{lbp}-processed image data is typically converted to histogram data and
used for training \gls{ml} models.
In~\cite{chingovska2012effectiveness}, the authors of the Replay-Attack dataset
perform experiments with an \gls{svm} that is trained with the \gls{lbp}
histogram data of greyscale images.
Boulkenafet~et~al.~\cite{boulkenafet2015face} extend this approach by applying
the \gls{lbp} feature on multiple color channels and by building the average of the
histograms of multiple consecutive frames.
Inspired by the \gls{lbp} operator, Zhao~et~al. introduce a novel \gls{cnn}
layer in~\cite{cdcn} that mimics the behavior of \gls{lbp} within neural
networks.

\section{Background}
\label{sec:background}

There exist multiple, highly diverse
solutions for liveness detection. Among other factors, the selection of sensors, considered attacks,
and facial features allows for various different approaches. In this section,
we cover general properties that fundamentally shape our approach and describe
three established datasets as well as performance metrics that we use for
evaluation.

\subsection{Sensors}
\label{sec:back-sensors}

Similar to facial recognition systems, liveness detection typically uses the
video data obtained from a camera that is pointed to the authenticating user.
On mobile devices, this provides images that consist of the three color
channels red, green, and blue (RGB). A limited share of devices has built-in
infrared sensors that can also be used for local liveness detection. Infrared
sensors help the system to obtain a \textit{depthmap} (three-dimensional depth
information) from the user's face in order to recognize typical face depth
profiles. Although this feature generally leads to better attack detection, it
greatly reduces the number of devices that are able to use the module. In our
work, we therefore focus exclusively on RGB cameras in order to obtain a model
that is compatible with all modern mobile phones.

\subsection{Attacks}
On a high level, each attack type exploits stolen data from a victim user
(e.g., a photograph of their face) and presents it to the liveness detection
module through the designated camera. In a \textit{print} attack, the attacker
presents a printout of the stolen photograph to the camera. Besides a
presentation of the flat printout, attacks are also possible with wrapped or
cut printouts. In more elaborate attacks, 3D (silicon) masks are formed from a
victim's photograph~\cite{khairnar2023face}.

For \textit{display} attacks, the attacker must obtain a video of the victim
user and present it via a computer display to the camera. Infrared sensors can
add another difficult hurdle, because attackers have to provide an authentic
three-dimensional depth profile of the victim. This can be achieved with
specially crafted plastic or silicon masks.

In our approach the focus lies on datasets that provide both printout and
display attacks.

\subsection{Photo and Video}

A discriminating factor in the field of liveness detection modules is given by
the number of images provided for each classification. We distinguish between
\textit{single-frame} (one image results in one classification) and
\textit{multi-frame} approaches (a number of consecutive frames is considered
for classification). In general, we notice that multi-frame approaches achieve
better classification accuracy and that the accuracy increases with a higher
number of consecutive frames. Limiting factors for the number of obtainable
consecutive frames are the user experience (authentication should be conducted
in real time) and the available data (users may not hold cameras steady over a
longer period of time).

\subsection{Color Spaces}
\label{sec:back-color}

\begin{figure}[t]
    \centering
    \includegraphics[width=0.66\linewidth]{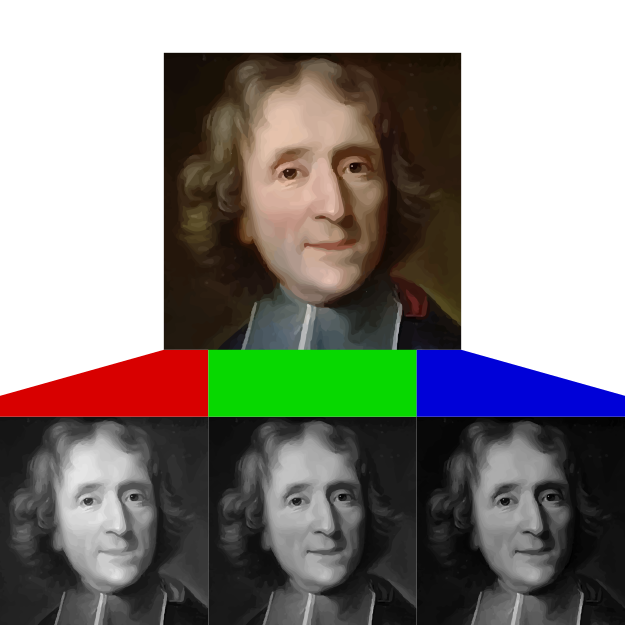}
    \caption{Original picture with RGB color channels: Red (left), Green
    (center), Blue (right).}
    \label{fig:rgb}
\end{figure}

\begin{figure}[t]
    \centering
    \includegraphics[width=0.30\linewidth]{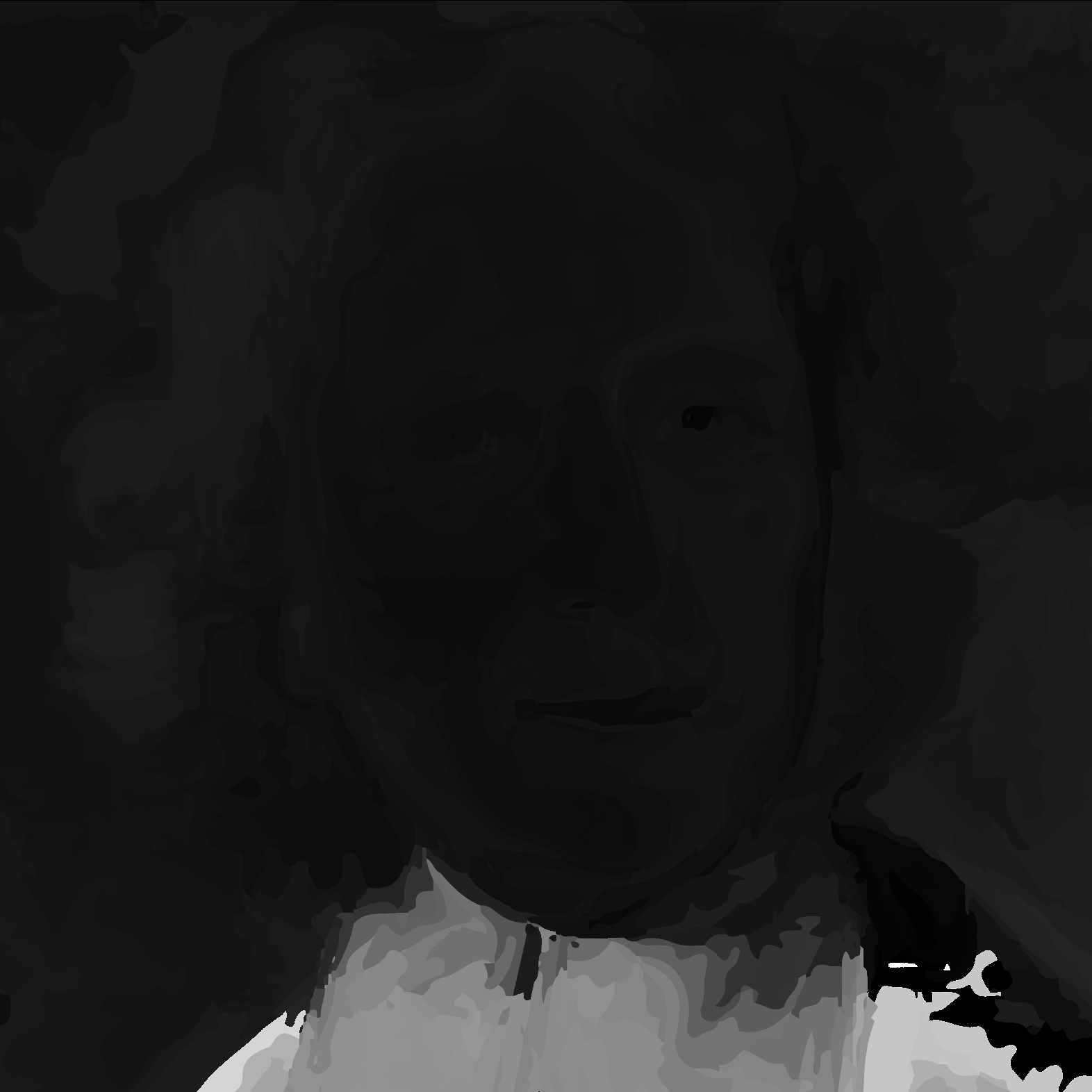}
    \includegraphics[width=0.30\linewidth]{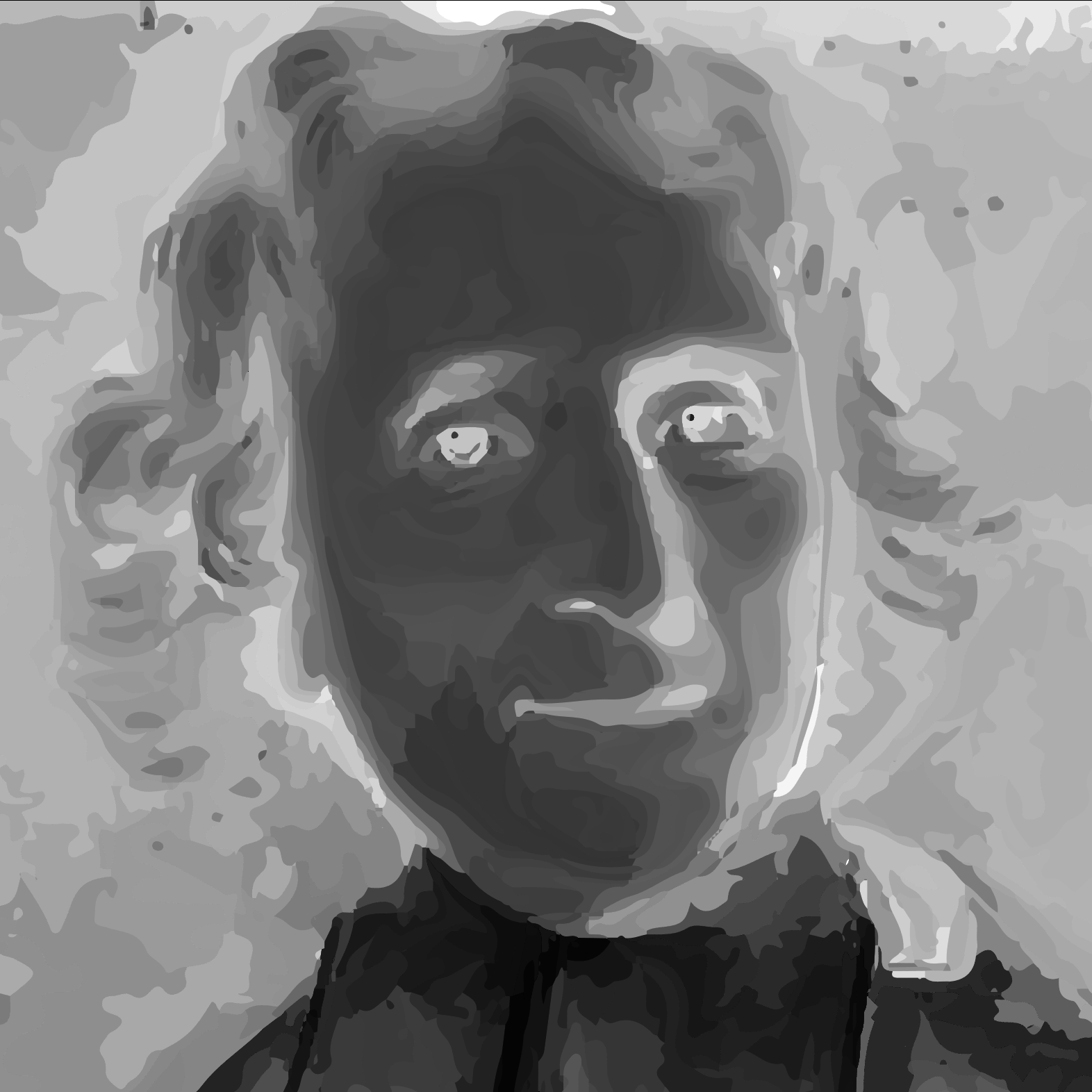}
    \includegraphics[width=0.30\linewidth]{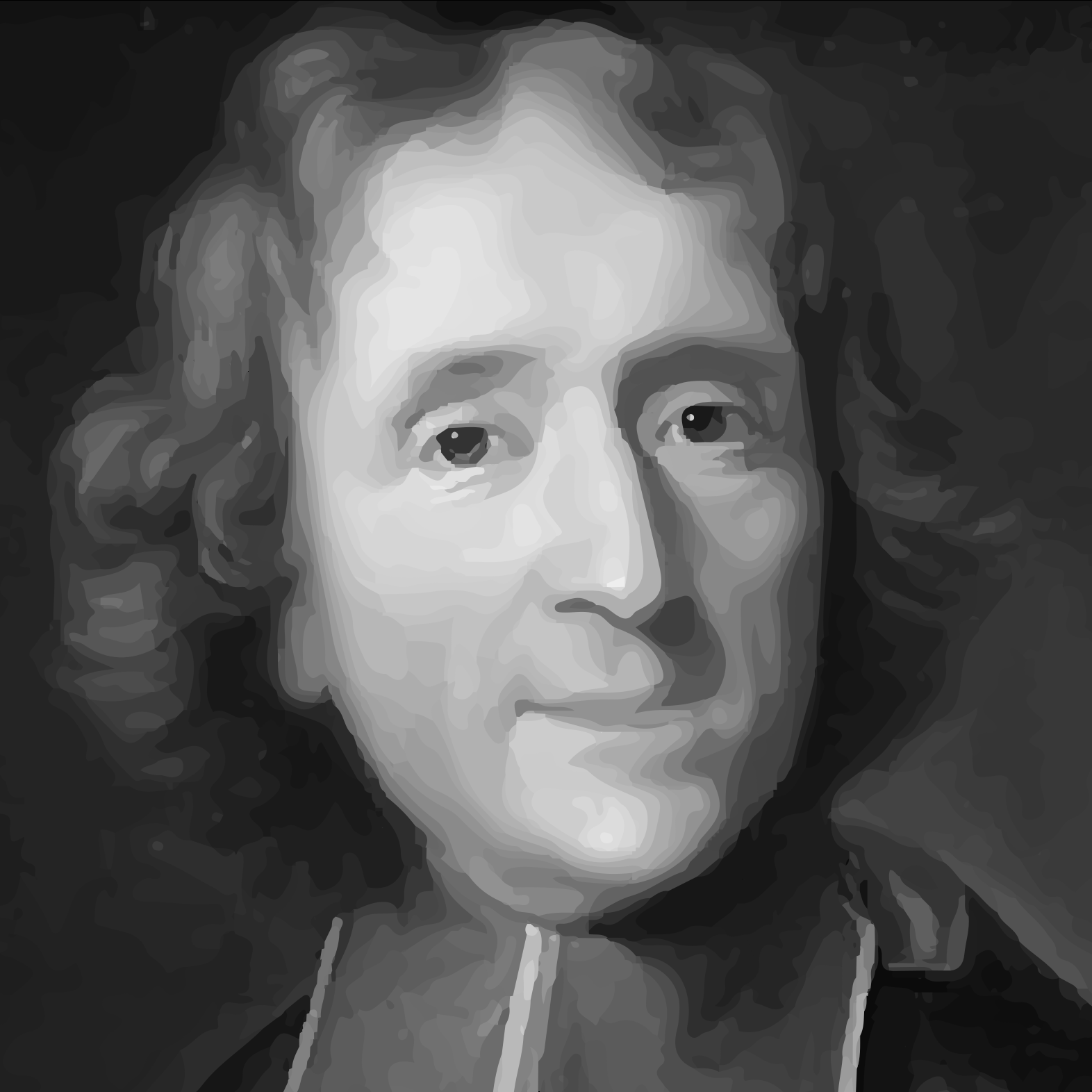}
    \includegraphics[width=0.30\linewidth]{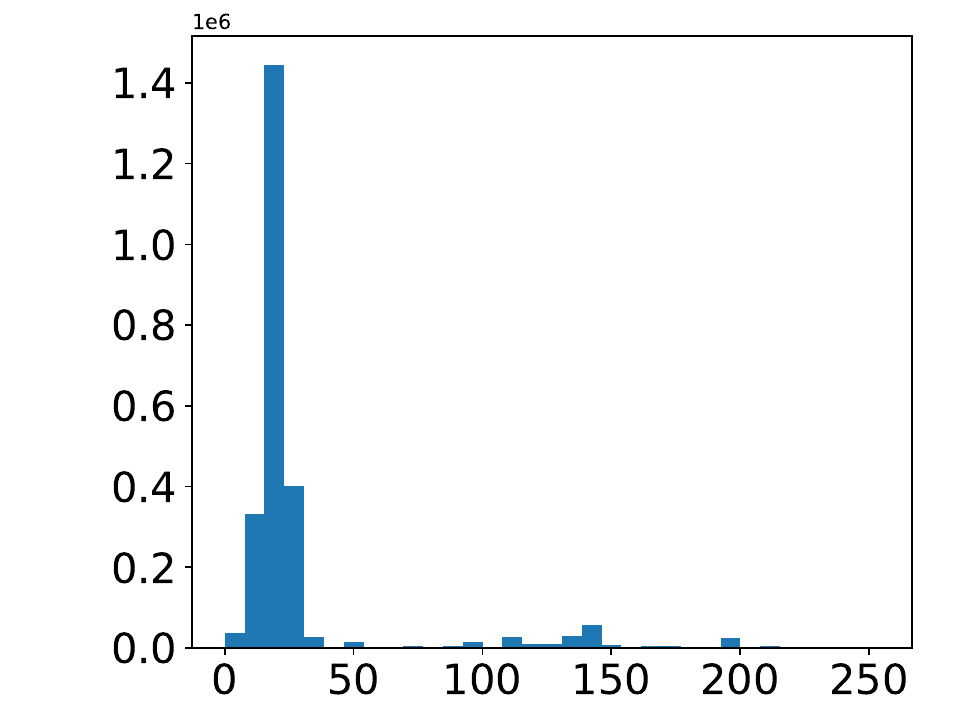}
    \includegraphics[width=0.30\linewidth]{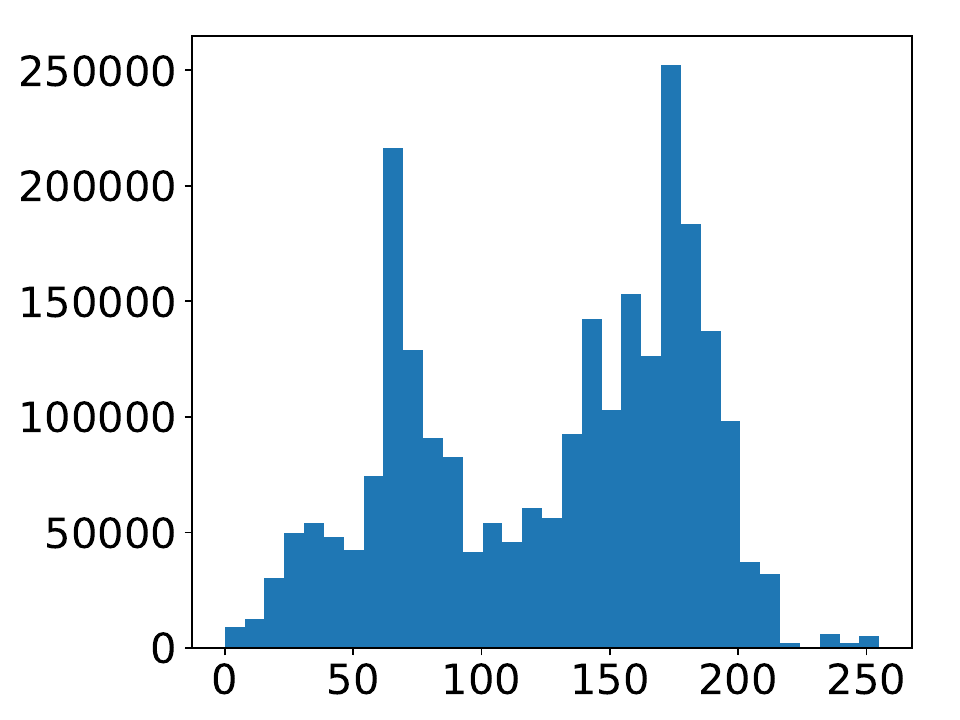}
    \includegraphics[width=0.30\linewidth]{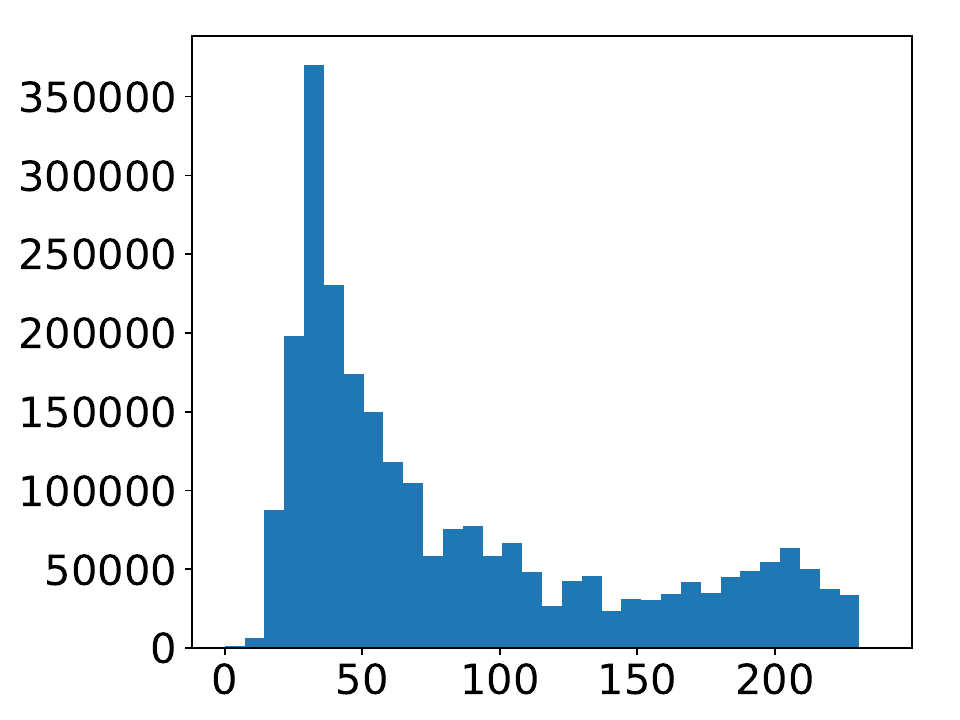}
    \caption{HSV color channels with corresponding histograms: Hue (left), Saturation (center), Value (right).}
    \label{fig:hsv}
\end{figure}

\begin{figure}[t]
    \centering
    \includegraphics[width=0.30\linewidth]{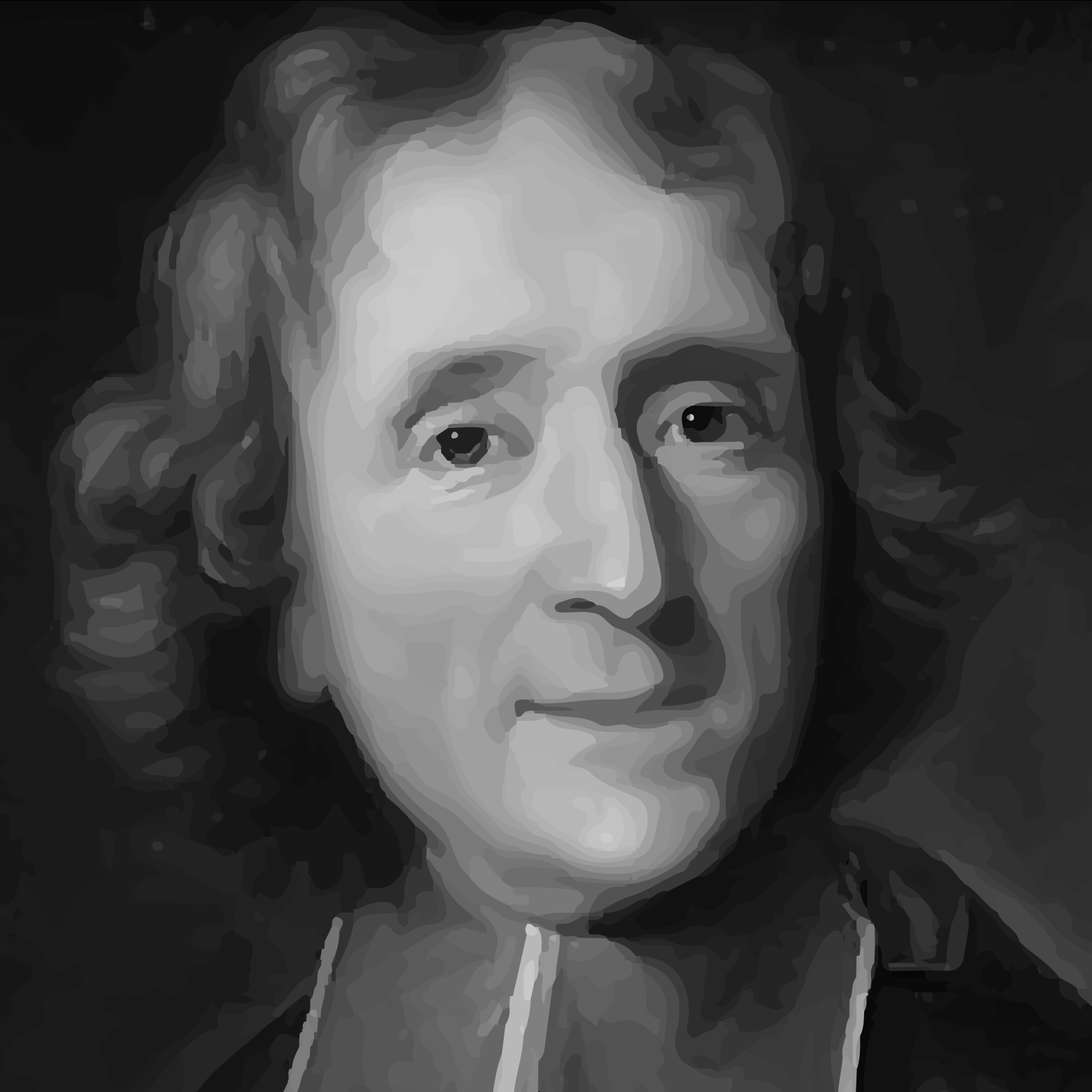}
    \includegraphics[width=0.30\linewidth]{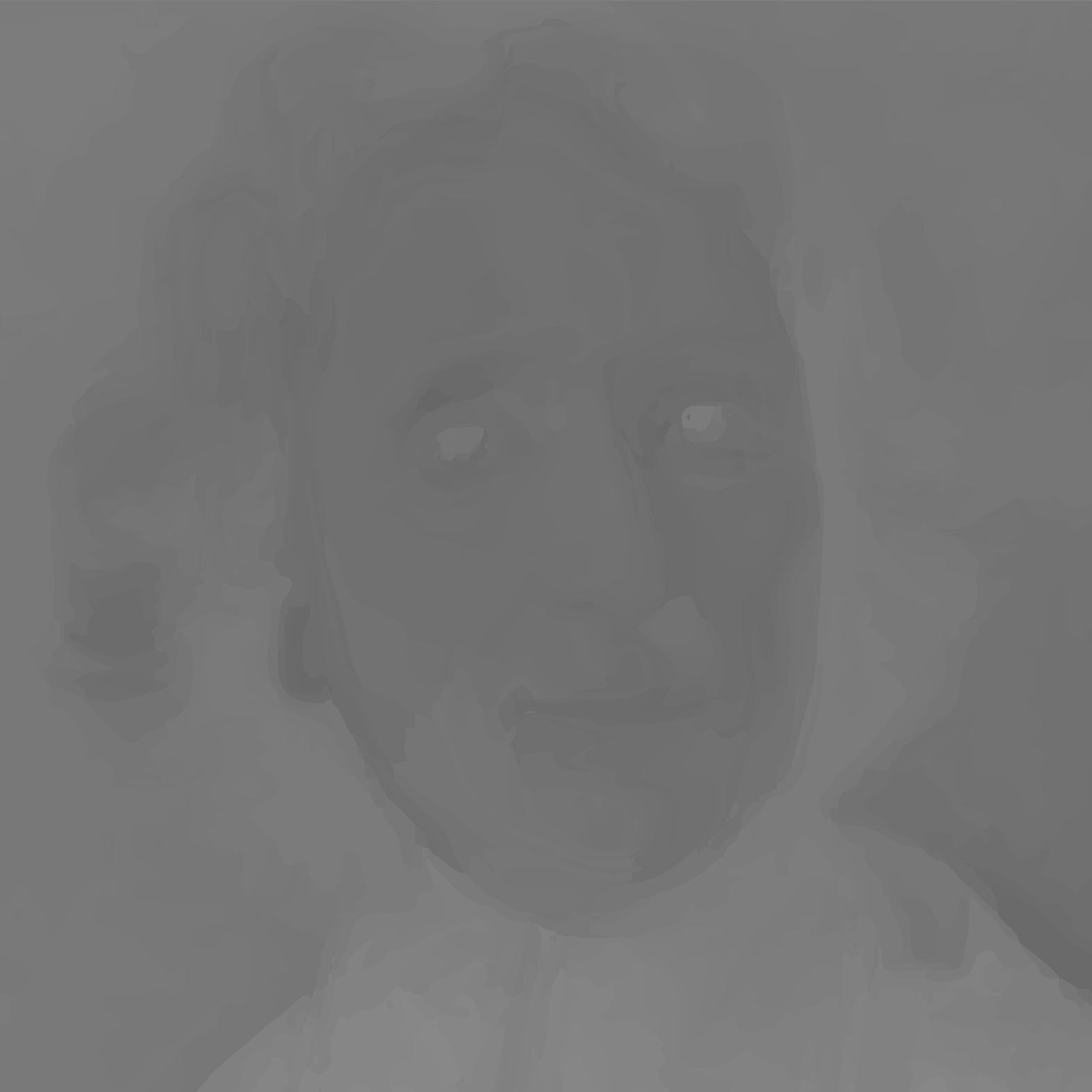}
    \includegraphics[width=0.30\linewidth]{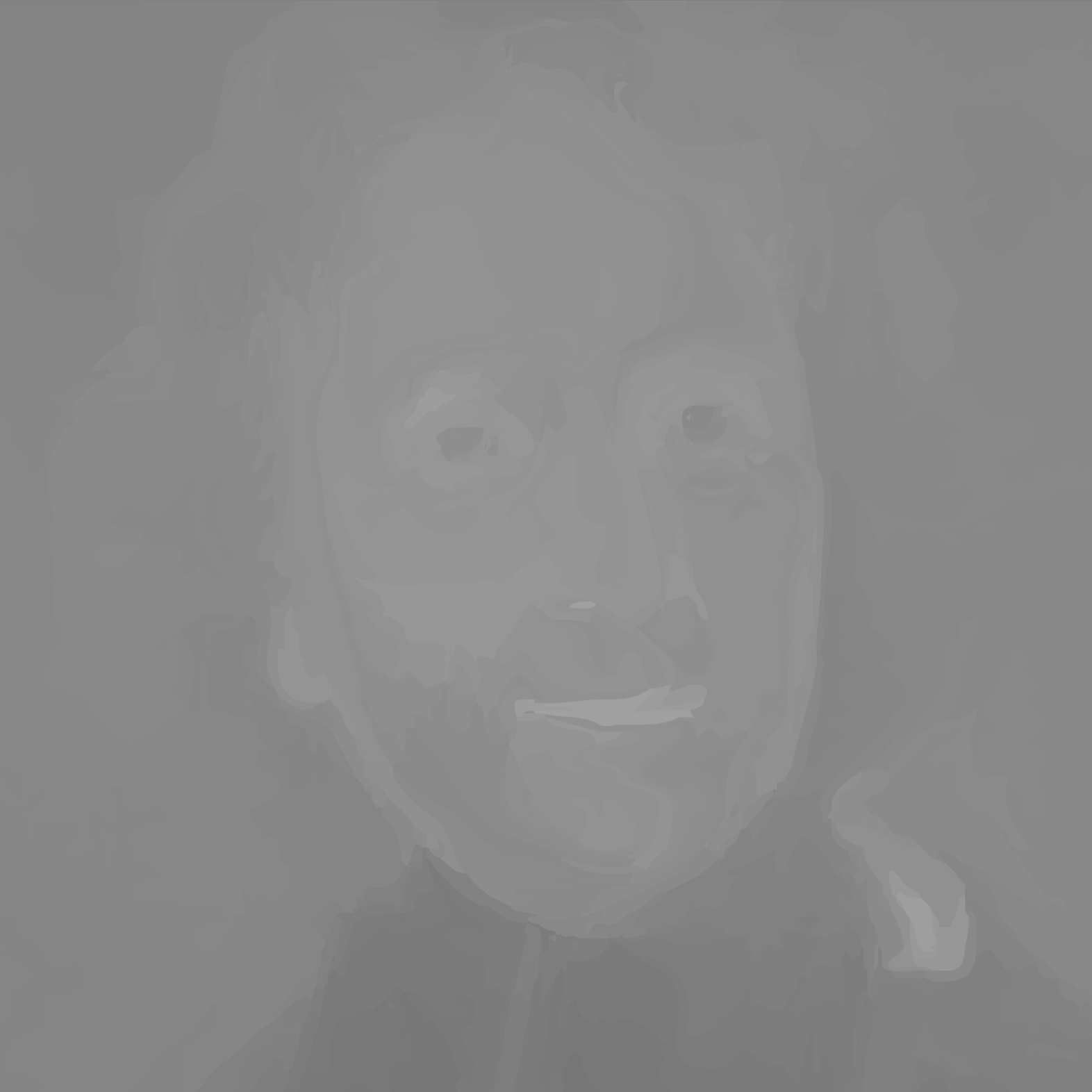}
    \includegraphics[width=0.30\linewidth]{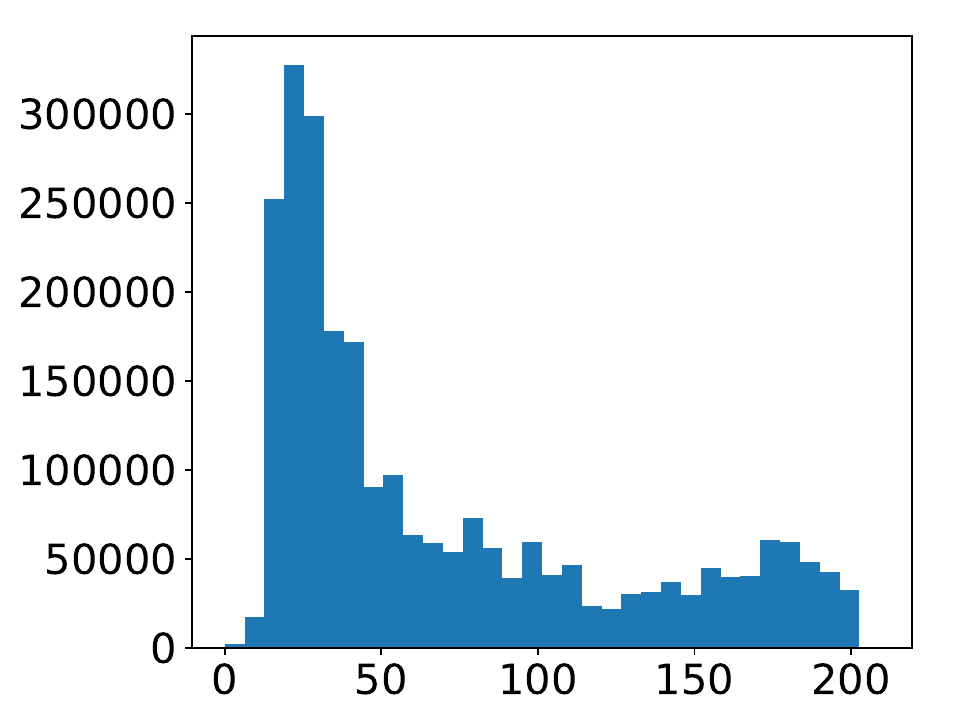}
    \includegraphics[width=0.30\linewidth]{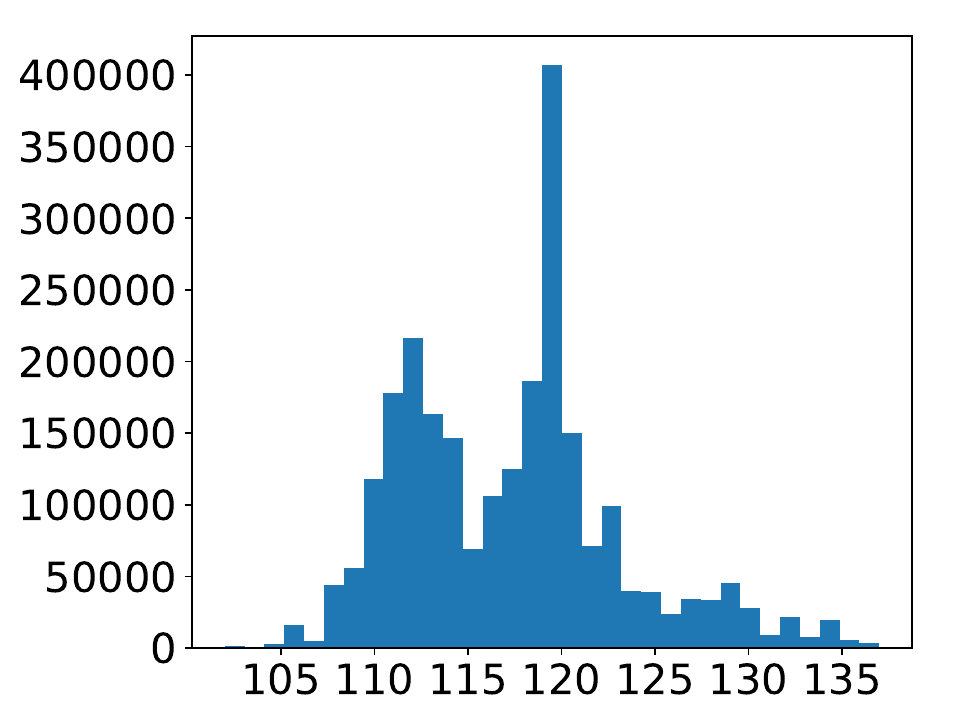}
    \includegraphics[width=0.30\linewidth]{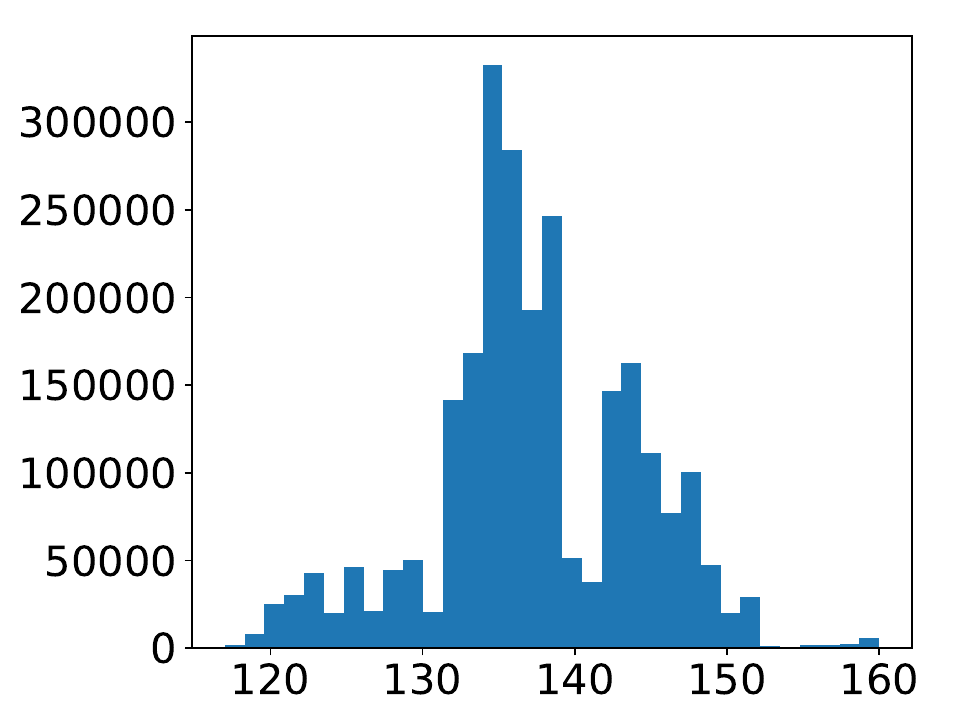}
    \caption{\ycbcr~color channels with corresponding histograms: Luma
    (left), Chrominance Blue (center), Chrominance Red (right).}
    \label{fig:ycbcr}
\end{figure}

As described in Section~\ref{sec:back-sensors}, the video data that is captured
with mobile devices is typically represented by the additive RGB color space that
consists of the three color channels red, green, and blue. As noted
in~\cite{boulkenafet2015face} and also observable in Figure~\ref{fig:rgb},
there is high correlation between these channels. Certain information such as
luminance and chrominance cannot be directly measured in RGB color space.

However, any image in RGB format can also be represented by color spaces such
as HSV and \ycbcr~that express more information. The HSV color space consists
of the three color channels hue, saturation, and value (see
Figure~\ref{fig:hsv}) where the latter represents luminance and the former two
express chrominance. Similarly, the \ycbcr~color space embodies the color
channels luma ($Y'$), chrominance red ($C_b$) and chrominance red ($C_r$) (see
Figure~\ref{fig:ycbcr}). Note that
in contrast to luminance that is obtained from \textit{linear} RGB color
channels, luma is based on RGB color channels that are
gamma-corrected~\cite{poynton2001yuv}. Both $C_b$ and $C_r$ express the
chromatic difference in relation to $Y'$.

Due to their characteristics, both HSV and \ycbcr~have already been used
for skin color detection and segmentation~\cite{shaik2015comparative} and
liveness detection~\cite{boulkenafet2015face}.

\subsection{Features}
\label{sec:back-features}

For the approach that is introduced with this paper we focus primarily on the
feature data obtained by the \gls{lbp} operator~\cite{ojala2002multiresolution}
and color histogram data. With this section, both utilized feature types are
described in detail.

\subsubsection{\gls{lbp}}

\begin{figure}[t]
    \centering
    \includegraphics[width=0.95\linewidth]{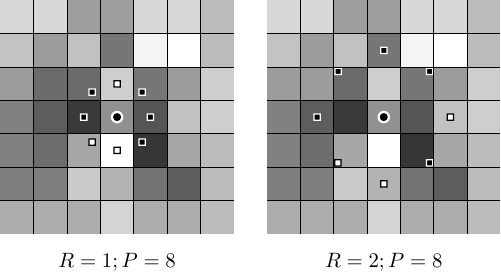}
    \caption{\gls{lbp} neighborhood of center pixel $c$ (marked with round disk) with $P=8$
    pixels for different radius values ($R$). The representation (black and
    white box) of pixel $p$ is filled black, if
    $\delta(g_p-g_c)=1$.}
    \label{fig:lbp-grid}
\end{figure}

\begin{figure}[t]
    \centering
    \includegraphics[width=0.95\linewidth]{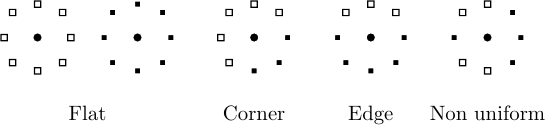}
    \caption{Typical patterns discovered with \gls{lbp}.}
    \label{fig:lbp-patterns}
\end{figure}

The \gls{lbp} operator, as proposed by Chan~et~al.~\cite{ojala2002multiresolution}, is
a special case of the Texture Unit that was initially developed by He and
Wang~\cite{he1990texture}. It is
extended for multiple color channels in~\cite{choi2010using} for the purpose of
face recognition. The operator has also been used for liveness detection for
its capability to describe the texture of images~\cite{boulkenafet2015face}.
However, in contrast to the existing work, we utilize an alteration
($LBP_{P,R}^{riu2}$) that is stable
against rotations in the plane~\cite{ojala2002multiresolution}.

Let $g_c$ be the gray value of a center pixel $c$ that has an evenly distributed
circular neighborhood of $P$ pixels where each pixel $p\in \{1,\dots,P\}$ has a distance of $R$
towards $c$ (see Figure~\ref{fig:lbp-grid}). Let
\begin{equation}
    \delta(x) =\left\{
        \begin{array}{ll}
            1, &x\geq 0\\
            0, &x<0
        \end{array}
        \right.
\end{equation}
such that $\delta(g_p - g_c)$ results in 0, iff the central pixel $c$ is brighter
than the pixel $p$.
The default \gls{lbp} operator
\begin{equation}
    LBP_{P,R}=\sum^{P-1}_{p=0}\delta(g_p-g_c)2^p
\end{equation}
uses this property to describe up to $2^P$
unique textures~\cite{ojala2002multiresolution}. However, this approach is not
well suited for the use case of liveness detection: it can be observed that by rotating the same image, different patterns are
produced. Hence, movement of the camera or user results in different patterns
for consecutive frames. We therefore utilize the rotation invariant uniform
pattern operator~\cite{ojala2002multiresolution}

\begin{equation}
    \label{eq:lbprriu2}
    LBP_{P,R}^{riu2} = \left\{
        \begin{array}{ll}
            \sum^{P-1}_{p=0}\delta(g_p-g_c), &\text{if }U(LBP_{P,R})\leq 2\\
            P+1, &\text{otherwise}
        \end{array}
        \right.
\end{equation}
with
\begin{align}
    U(LBP_{P,R}) &= | \delta(g_{P-1}-g_c) -\delta(g_0-g_c)|\\
    &+ \sum^{P-1}_{p=1}|
    \delta(g_{p}-g_c) -\delta(g_{p-1}-g_c)|.
\end{align}
$U$ represents the number of transitions (the number of events where the value
of $\delta$ changes) along the circular neighborhood. In the case of the flat,
corner, and edge patterns illustrated in Figure~\ref{fig:lbp-patterns}, $U$
produces a value of $\leq 2$ such that the patterns are considered \textit{uniform}.
In contrast, $U$ produces a value of $4$ for the \textit{non uniform} pattern
in Figure~\ref{fig:lbp-patterns}. From Equation~\ref{eq:lbprriu2} it can be observed that $LBP_{P,R}^{riu2}$ does not
distinguish between non-uniform samples. Instead, all such samples are assigned
to the value $P+1$. Furthermore, the output dimension of
$LBP_{P,R}^{riu2}$ is greatly reduced ($P+2$ patterns) in comparison to $LBP_{P,R}$ that
produces up to $2^P$ patterns.

\begin{figure}[t]
    \centering
    \includegraphics[width=0.30\linewidth]{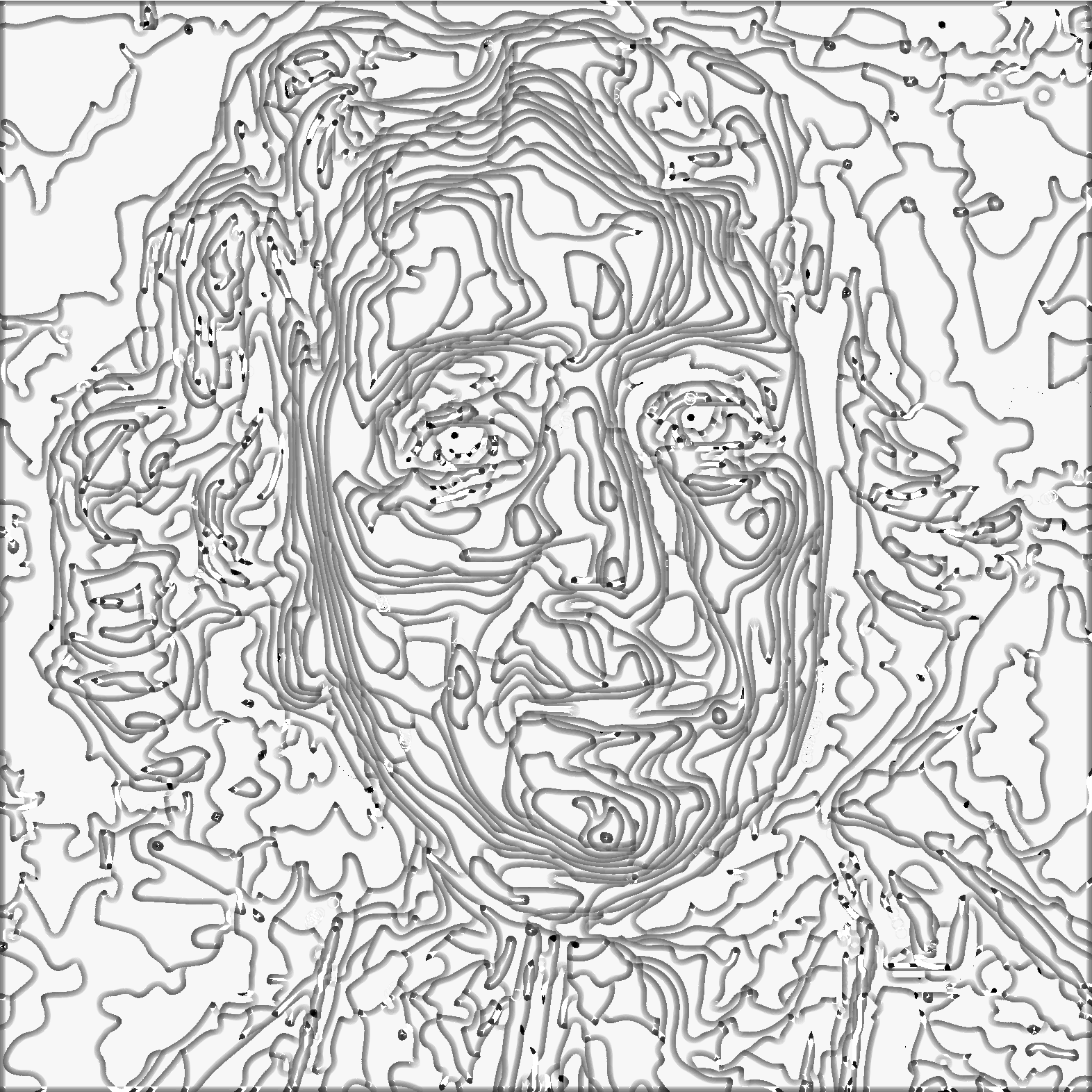}
    \includegraphics[width=0.30\linewidth]{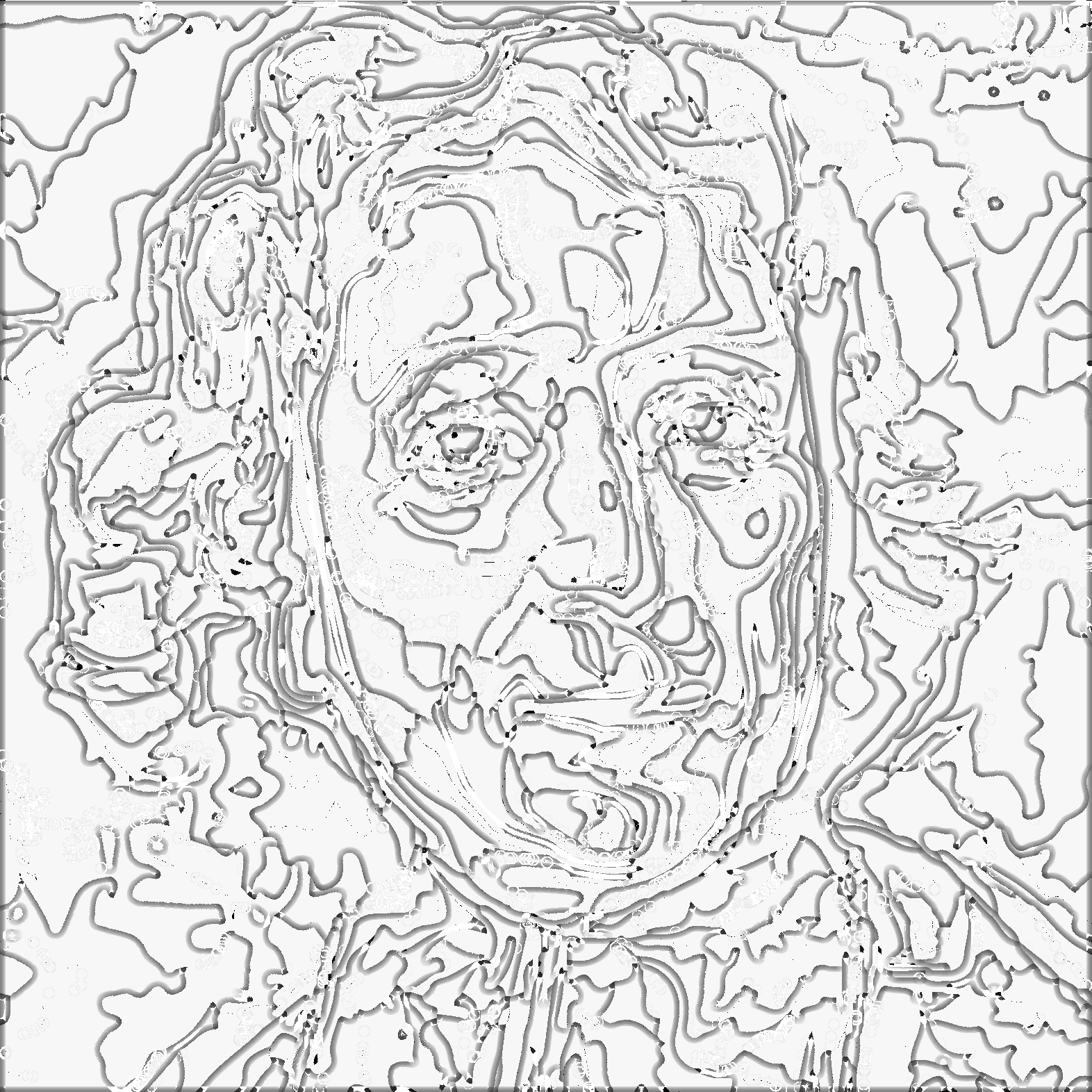}
    \includegraphics[width=0.30\linewidth]{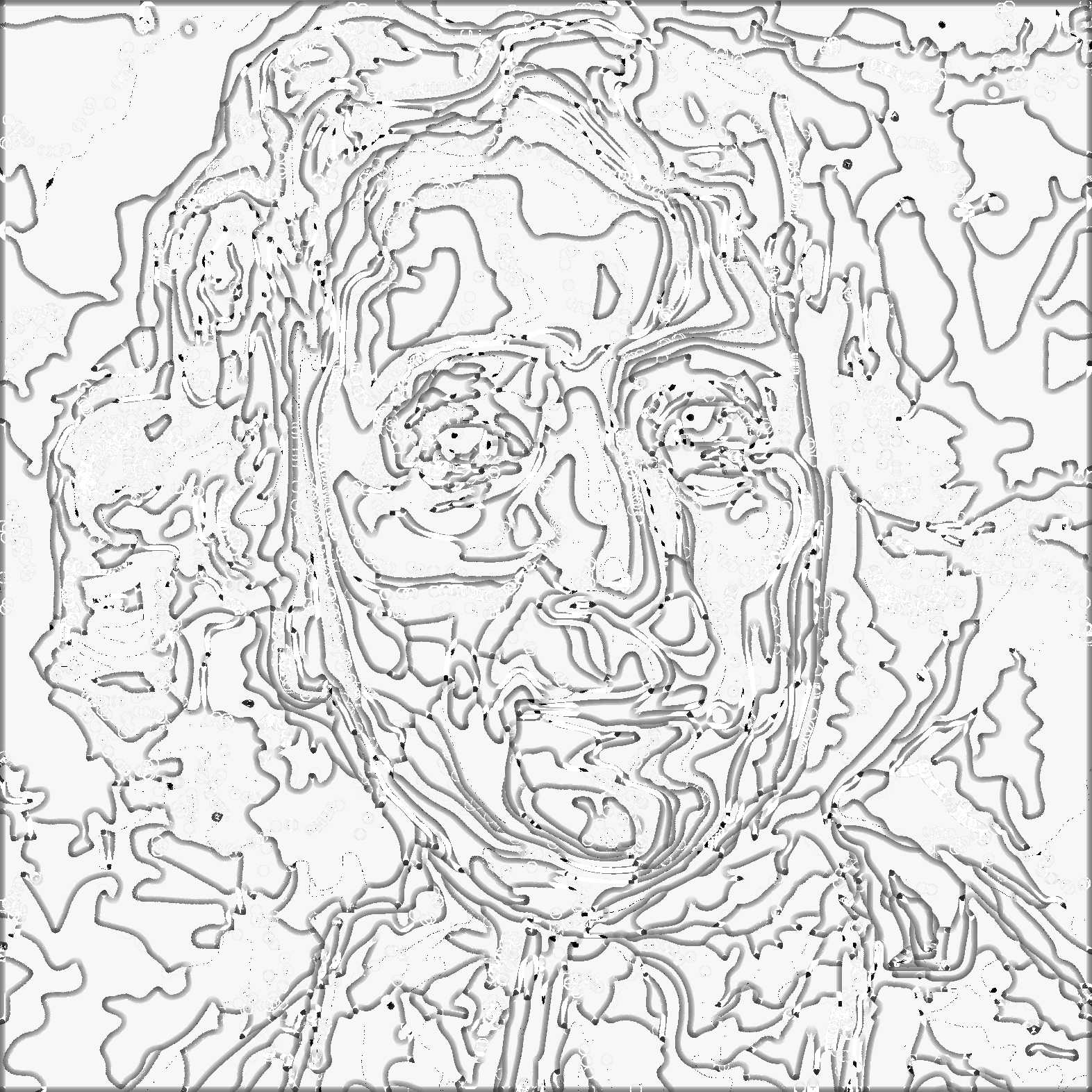}
    \includegraphics[width=0.32\linewidth]{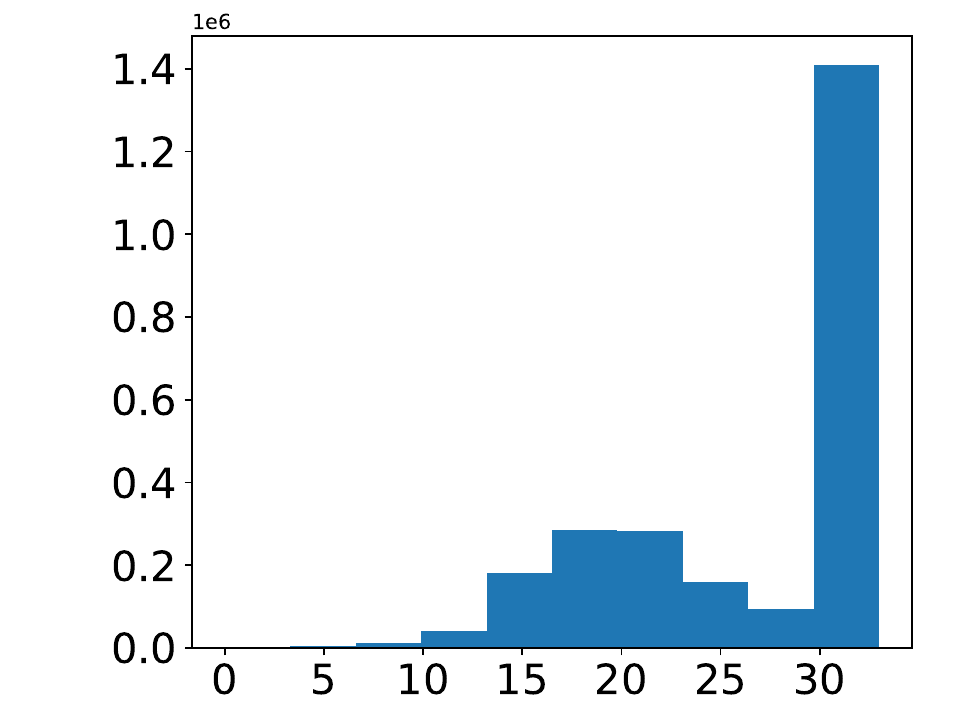}
    \includegraphics[width=0.32\linewidth]{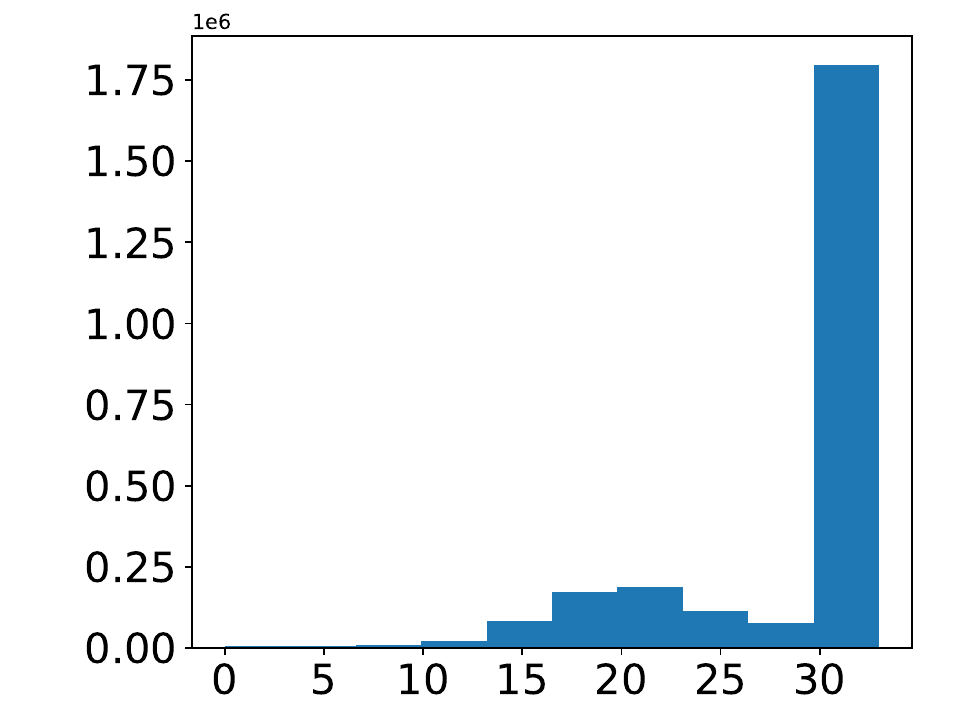}
    \includegraphics[width=0.32\linewidth]{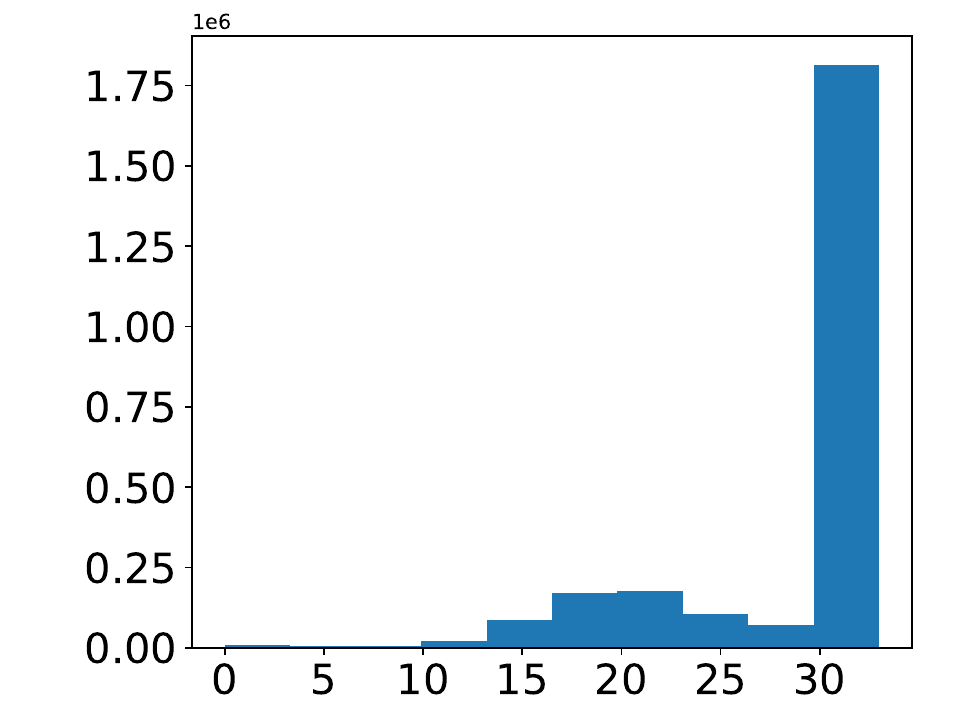}
    \caption{\ycbcr~color channels (with corresponding histograms) processed with
    the rotation invariant uniform pattern $LBP_{P,R}^{riu2}$ for $P=32, R=8$.}
    \label{fig:ycbcr-lbp}
\end{figure}

Our approach does not use the $LBP_{P,R}^{riu2}$-processed frames directly.
Instead, similar to~\cite{choi2010using} and~\cite{boulkenafet2015face}, the
operator is individually applied on a set of color channels. The processed data
of each color channel is then translated to histogram representation. Finally,
the histograms of all color channels are concatenated, forming our feature
representation for single frames.

\subsubsection{Color Histogram}
We observe that in many
cases, existing work on liveness detection exclusively relies on the
textural information that is obtained with the \gls{lbp} feature. However, it
must be noted that \gls{lbp} does not preserve the overall color information.
We therefore choose to also utilize the direct histogram representations of
each frame's
color channels as feature data. 

\subsection{\gls{lstm} Layers}

This work proposes a \gls{ml} model that utilizes \gls{lstm}
layers~\cite{hochreiter1997long}. \gls{lstm} is a special type of \glspl{rnn}.
The method eliminates the issue of decaying error backpropagation in classical
neural networks through a combination of \textit{memory cells} and \textit{gate
units} that regulate which information is stored and/or
forgotten~\cite{hochreiter1997long}. As such, \gls{lstm} layers prove to be
well suited for predicting and classifying time series
data~\cite{van2020review}. They were utilized for liveness detection
before~\cite{tu2019enhance}. In this work, \gls{lstm} layers are used to learn
the temporal information of feature data from consecutive video frames.

\subsection{Datasets}

For liveness detection experiments, models are typically trained with
pre-published datasets. A list of available datasets is compiled in~\cite{dataset_survey} and
covers a multitude of datasets of photos and videos recorded with different
sensors and various attack scenarios in mind. We limit ourselves to datasets
that were recorded with ordinary mobile phone cameras and contain photos and
videos of bonafide access attempts and replay attack attempts via paper
printouts and computer displays.

For our experiments, we focus on the datasets described below. They all have in
common that for every participating user, \textit{live videos} were recorded
with various cameras, before those recordings were used to simulate
\textit{attack attempts}. For print attacks, videos were recorded by a hidden
person that holds a randomly selected printed out video frame in front of the
camera. Replay attacks were recorded by a hidden person that displays either a
single frame or an entire video on a digital screen (e.g. smartphone display).
Every dataset therefore consists entirely of videos. These videos were recorded
in multiple sessions in which various parameters such as attack type, replay
device, lighting conditions, user behavior, etc. were altered. A single session
is considered a subset of the dataset containing videos from all users with
certain parameters being fixed. All datasets, except for RECOD-MAPD, define a
split of the userbase into two or three subsets \textit{training},
\textit{validation}, and \textit{test}. We closely follow this split for the
evaluation of our approaches.

\subsubsection{OULU-NPU}

OULU-NPU~\cite{boulkenafet2017oulu} contains 4950 videos of 55 users recorded
with six different phone cameras of varying price range and quality. The
dataset is split into 1800 training and testing videos respectively. 1350
videos are reserved for validation (see Table~\ref{table:oulu-npu}). Attack
presentations include both printouts and video playback. 

\begin{table}[h!]
\centering
\caption{OULU-NPU Dataset. Users are split by Training, Validation, and Test}
\begin{tabular}{|r||c|c|c|c|c|}
\hline
 Split & No. Users & Live & Print attacks & Display attacks & Total  \\
 \hline
 \hline
 Training & 20 & 360 & 720 & 720 & 1800  \\
 \hline
 Validation & 15 & 270 & 540 & 540 & 1350  \\
 \hline
 Testing & 20 & 360 & 720 & 720 & 1800 \\
 \hline
\end{tabular}
\label{table:oulu-npu}
\end{table}

\subsubsection{RECOD-MPAD}

RECOD-MPAD, published in~\cite{almeida2020detecting}, is a dataset that
contains 2250 videos of 45 users that come preprocessed as single frames. The
videos were taken with two different phones and five different lighting
scenarios, including outdoor recordings. The attack scenarios contain two
printout attacks (different lighting) and two screen replay attacks (TV screen
and computer monitor). The dataset is not split by the authors. We therefore
select a manual split as follows: users 1 to 17 are used for training, users 18
to 30 are used for validation, and users 31 to 45 are used for testing. This
results in a distribution of videos as listed in Table~\ref{table:recod-mpad}.

\begin{table}[h!]
\centering
\caption{RECOD-MPAD Dataset with custom user split.}
\begin{tabular}{|r||c|c|c|c|c|}
\hline
 Split & No. Users & Live & Print attacks & Display attacks & Total  \\
 \hline
 \hline
 Training & 17 & 170 & 340 & 340 & 850  \\
 \hline
 Validation & 13 & 130 & 260 & 260 & 650  \\
 \hline
 Testing & 15 & 150 & 300 & 300 & 750 \\
 \hline
\end{tabular}
\label{table:recod-mpad}
\end{table}

\subsubsection{SiW}
The authors of~\cite{liu2018learning} published SiW, a large dataset that
contains videos of 165 users. Videos are recorded in 1080p at 30 frames per
second with a DLSR camera and a high quality webcam. For each user, eight
videos with different poses, distances and lighting conditions are recorded.
The attack attempts include up to 20 printout and display replay attacks per
user. In total, there are 4462 videos in the dataset (see Table~\ref{table:siw}).

\begin{table}[h!]
\centering
\caption{SIW Dataset. Users are split by Training and Test}
\begin{tabular}{|l||l|l|l|l|l|}
\hline
 Split & No. Users & Live & Print attacks & Display attacks & Total  \\
 \hline
 \hline
 Training & 90 & 714 & 339 & 1364 & 2417  \\
 \hline
 Testing & 75 & 583 & 314 & 1148 & 2045 \\
 \hline
\end{tabular}
\label{table:siw}
\end{table}

\subsubsection{Evaluation Protocols}

For OULU-NPU and SiW the authors also provide evaluation protocols. These are
subsets of the entire dataset that hold certain properties. For example, in
some of our experiments we limit ourselves to \textit{OULU protocol I}. In that
case, the model is trained only on sessions 1 and 2 and tested on session 3
(which introduces a new room lighting that was not seen in the former
sessions).
Note that
in all of our experiments the user base for training, evaluation, and testing
is completely disjunct.

\subsubsection{Cross Dataset Testing}

In addition to using dataset specific evaluation protocols, the
generalizability of a liveness detection model can also be assessed with
\textit{cross dataset testing}. With this technique, two entirely different
datasets are used. The model is trained with the videos of one dataset and
tested with the videos of the other dataset. This requires the model to perform
well under entirely unseen conditions (different lighting, different cameras,
different camera poses, different user groups, etc.).

\subsection{Metrics}

In the field of anti-spoofing models, the performance of detection modules is
measured by different types of error rates~\cite{khairnar2023face}:

\noindent\textbf{BPCER:} The Bona Fide Presentation Classification Error Rate
(BPCER) expresses the rate at which authentication attempts of benign users are
mistakenly classified as an attack (i.e., benign users cannot access the
service).

\noindent\textbf{APCER:} The Attack Presentation Classification Error Rate
(APCER) expresses the rate at which attack attempts of malicious users are
mistakenly classified as benign (i.e., attackers obtain access to a restricted
service).

\noindent\textbf{ACER:} This metric evaluates the mean of APCER and BPCER
($\text{ACER}=\frac{\text{APCER} + \text{BPCER}}{2}$) and is commonly used to
express the overall performance.

\noindent\textbf{HTER:} The Half Total Error Rate (HTER) represents the mean of
the FAR (False Positive Rate) and the FNR (False Negative Rate)
($\text{HTER}=\frac{\text{FPR}+\text{FNR}}{2}$) and is used to measure
performance across datasets~\cite{khairnar2023face}.

\noindent\textbf{Balanced Accuracy:} In addition to error rates, the model
performance is also expressed with an accuracy metric. Because of a high
disparity in benign and attack samples, it is required to balance the metric.
Balanced Accuracy is defined as $\frac{TPR + TNR}{2}$ where TPR (True Positive
Rate) corresponds to the rate at which benign users are correctly accepted and
where TNR (True Negative Rate) corresponds to the rate at which attacks are
correctly detected.

\section{Model Design}
\label{sec:model-design}

This section describes the design of our approach, termed \gls{ctlstm} for liveness detection. It
emphasizes the importance of feature engineering and trains a \gls{ml} model
with an abstract representation of the original video data.

\subsection{Data Preparation}

As described in Section~\ref{sec:back-features}, \gls{lbp} and color channel
histograms are both useful features for liveness detection. We choose to
concatenate both features and use those as input for a \gls{lstm} \gls{ml}
architecture. \gls{lstm} networks are known to work well on time series data.
We make use of this property by collecting feature data of multiple consecutive
frames in each training sample. As a result, each sample can be viewed as a
short time series of histograms.

In order to obtain a time series of histograms, the videos of all available
datasets are split into separate samples with a constant number of frames ($n$)
per sample. Each frame is transformed into the \ycbcr~and HSV color spaces. The
color channels of each color space are then processed separately by the
\gls{lbp} and the color histogram feature. Note that the final step of the
\gls{lbp} feature also computes the histogram data of the \gls{lbp} filtered
frame. We therefore obtain two histograms per color channel (one from the
\gls{lbp} feature and one from the color histogram feature). For the input
data, we have the following tunable parameters:

\begin{itemize}
    \item Number of frames per sample ($n$).
    \item Number of histogram buckets ($m$): histograms count the number of
        occurrences of values within certain value ranges. Each bucket
        represents an individual value range. By increasing $m$, the ranges
        become more fine-grained. An oversized $m$ can lead to too much detail
        where useful information can no longer be obtained. An undersized $m$
        can result in critical information being missed by the \gls{lstm}. We
        define $m_1$ as the number of buckets for the color histogram feature
        and $m_2$ as the number of buckets for the \gls{lbp} feature.
    \item \gls{lbp} radius ($R$): the radius $R$ determines the distance of sample points to the center pixel. Consequently, all sample points are located on a circle around the center pixel.
    \item Number of \gls{lbp} points ($P$): $P$ denotes the number of sample points that are considered for each center pixel. Every sample point has a distance of $R$ to the center pixel.
    \item Number of color channels ($c$): in the case that only \ycbcr~or HSV is used, $c$ equals $3$. However, the model allows that multiple color spaces are used. In the case that both \ycbcr~and HSV are used, $c$ equals $6$.
\end{itemize}

As described earlier, the histogram values of all features are concatenated for
each frame. Hence, each frame is represented by a vector $v$ with a length of
$c\cdot (m_1+m_2)$. A sample with $n$ frames equals a matrix of dimension
$n\times (c\cdot (m_1+m_2))$.

The raw data that is obtained with the methods described above must be
normalized before it can be passed to the \gls{lstm} model. For this purpose,
z-score normalization is applied on the feature data of all datasets as
follows. Let
\begin{equation}
    v_i' = \frac{v_i-\mu_i}{\sigma_i},
\end{equation}
then $v'$ is the normalized version of feature vector $v$, $\mu_i$ represents
the mean and $\sigma_i$ represents the standard deviation of the $i$th feature
value. The values of $\mu_i$ and $\sigma_i$ are collected from the training
part of each dataset and reused for their respective validation and testing
parts.

\subsection{Network Architecture}

\begin{table}
\centering
\caption{\gls{lstm} Network Architecture.}
\begin{tabular}{| c | l | c | c |} 
 \hline
 \# & Layer & Input Size & Output Size \\ [0.5ex] 
 \hline\hline
 1 & Dropout ($p=10\%$) & $c\cdot (m_1+ m_2)$ & $c\cdot (m_1+ m_2)$ \\ 
 2 & \gls{lstm} & $c \cdot (m_1+ m_2)$ & 480 \\
 3 & Dropout ($p=50\%$) & 480 & 480 \\
 4 & Linear & 480 &  2\\
 \hline
\end{tabular}
\label{table:lstm}
\end{table}
\begin{table}
\centering
\caption{Dual-Layer \gls{lstm} Network Architecture.}
\begin{tabular}{| c | l | c | c |} 
\hline
\# & Layer & Input Size & Output Size \\ [0.5ex] 
\hline\hline
1 & \gls{lstm} & $c \cdot (m_1+ m_2)$ & 240 \\
2 & Dropout ($p=40\%$) & 240 & 240 \\
3 & \gls{lstm} & 240 & 240 \\
4 & Linear & 240 &  240\\
5 & ReLU & 240 &  240\\
6 & Linear & 240 &  2 \\
7 & Log. Softmax & 2 &  2 \\
\hline
\end{tabular}
\label{table:dual-lstm}
\end{table}

The sequential \gls{ml} architecture is listed in Table~\ref{table:lstm}: a
single \gls{lstm} layer, as defined in~\cite{hochreiter1997long}, learns the
temporal color and texture features. The \gls{lstm} layer is surrounded by
dropout layers that aim to prevent model overfitting. At each dropout layer, a
relative proportion of features ($p$) is randomly removed from training. The
final layer is a linear densely-connected layer that reduces the output size to
two values that can be used for binary classification. For performance comparison,
an additional dual-layer \gls{lstm} model is developed that utilizes two
\gls{lstm} layers and a larger set of linear and rectified linear unit layers (see Table~\ref{table:dual-lstm}).

The network uses the optimization algorithm RMSprop (root mean square
propagation)~\cite{hinton2012neural} that is based on a moving average of the
square of gradients. With the results of a grid search it was concluded that
this algorithm has better performance than other known algorithms such as Adam
and Stochastic Gradient Descent (SGD).

\subsection{Hyperparameter Grid Search}
\label{sec:hyper-grid-search}

A hyperparameter grid search can be used for determining the parameter values
that allow maximal performance. For this work, we perform an early grid search
for finding the initial coarse parameters that would be fine-tuned manually
later on. The grid search is also used for comparing the single-layer and
dual-layer \gls{lstm} approaches. All utilized grid search parameters are
listed in Table~\ref{table:lstm-grid-search}: we apply different optimization
algorithms, batch sizes, dropout frequencies, and different numbers of hidden
units. For this grid search, the feature parameters are constant. The following
values are being used: color histogram buckets: 50; \gls{lbp} histogram buckets: 9;
\gls{lbp} sample points: $P=8$; \gls{lbp} radius: $R=8$; color space: \ycbcr; frames per
sample: 10.

The search result shows that

\begin{itemize}
    \item the best performance is achieved with 240 to 480 hidden units,
    \item the best dropout frequency for the first and second dropout layers are 0.4 to 0.5 and 0.1 respectively,
    \item RMSprop proves to be the best suited optimizer,
    \item both single-layer and dual-layer \gls{lstm} models have similar performance, and
    \item the batch size has no significant performance impact. We continue to use a batch size of 32.
\end{itemize}

\begin{table}
\centering
\caption{Grid Search Parameters for (dual-layer) \gls{lstm} tuning.}
\begin{tabular}{| l | c | c | c |} 
 \hline
 Optimizer & Batch Size & Dropout & Hidden Units \\ [0.5ex] 
 \hline\hline
 SGD     & 32 & 0.075 & 64 \\
 Adam    & 48 & 0.125 & 128 \\
 RMSprop & 64 & 0.2   & 240 \\
         & 96 & 0.5   & 480 \\
 \hline
\end{tabular}
\label{table:lstm-grid-search}
\end{table}

\section{Evaluation}
\label{sec:evaluation}

With this section we describe our experiments and
the achieved performance results.
The proposed \gls{ctlstm} model is evaluated on the datasets
OULU-NPU~\cite{boulkenafet2017oulu}, RECOD~\cite{almeida2020detecting}, and
SiW~\cite{liu2018learning}.

For evaluation we take all \gls{lbp}, color histogram, and \gls{ml} related
parameters into consideration. Following the grid search described in
Section~\ref{sec:hyper-grid-search}, we focus on the tuning of feature parameters and
perform tests in different dataset constellations by following certain dataset
protocols and performing cross-dataset training.

In all experiments (except for the cross-dataset test), the dataset utilization
is the same: during training, the model sees the train subsets of either a single dataset
or all three datasets. The performance on the validation and test subsets is reported for all
three datasets individually.

\subsection{Feature Parameter Tuning}

\begin{table}[t]
\centering
\caption{Assessed feature parameter values. The values that are used by the
    proposed model are in bold.}
\begin{tabular}{| l | l |} 
 \hline
 Parameter & Values \\
 \hline\hline
    Frames per Sample ($n$) & 10 / \textbf{16} / 30 / 60 \\
 \hline
    Color Space & HSV / \ycbcr / \textbf{HSV AND \ycbcr} \\ 
 \hline
    Color Histogram Buckets ($m_1$) & 8 / 32 / \textbf{50} / 64 \\
 \hline
    \gls{lbp} Hist. Buckets ($m_2$) & 10 / \textbf{34} / 66 \\
 \hline
    \gls{lbp} Sample Points ($P$) & 8 / \textbf{32} / 64 \\
 \hline
    \gls{lbp} Radius ($R$) & 1 / \textbf{8} \\
 \hline
\end{tabular}
\label{table:lstm-parameter-set}
\end{table}

\subsubsection{\gls{lbp} \& Histogram Tuning}
The initial tuning of the \gls{lbp} feature is performed with a graphical
program that applies the \gls{lbp} filter in real time on any specified video.
The program's source code can be obtained from the authors of this paper. Via
the program's visual feedback it is determined that an \gls{lbp} radius of
$R=8$
produces the richest texture information. This observation is contrary to
values used in existing work; e.g., \cite{boulkenafet2015face} use a radius of
$R=1$. However, a dedicated experiment with our model shows that $R=8$
achieves considerably better performance.

The complete set of assessed parameter values is listed in
Table~\ref{table:lstm-parameter-set}. For the number of \gls{lbp} sample points
it was determined that a larger number (e.g., $P=32$) results in better
performance. However, with even larger values (such as $P=64$) it is observed that
computation time becomes infeasible. We therefore choose to set this value to
$P=32$. Contrary to
the classic $LBP$ operator, the $LBP_{P,R}^{riu2}$ operator that is used in
this work only produces $P+2$ different values. Therefore, the number of
histogram buckets for the \gls{lbp} feature can be restricted to $34$.

In the case of the color histogram feature, the most important controllable
parameter is the number of histogram buckets. It is found that a small number
of buckets leads to information loss, while 50 to 64
buckets capture a beneficial level of detail. For evaluation, we choose 50
buckets for the color histogram feature.

\subsubsection{Color Space Selection}
In a first test, we perform feature preparation for different color spaces
(HSV, RGB, \ycbcr) separately. Here, the best performance is achieved with \ycbcr.
However, subsequent tests with features applied on both HSV and \ycbcr~reveal a
considerable performance boost. For the same configuration as in the first
experiment, a \gls{ctlstm} model that uses both HSV and
\ycbcr~feature data achieves an ACER of 0.0897 on the validation data of
OULU-NPU whereas an
\gls{ctlstm} trained only with \ycbcr~feature data achieved an ACER of 0.1944. The
full list of achieved values is provided with
Table~\ref{table:lstm-color-space-utilization}. It must be noted that the training for
these experiments covered only 17 epochs with non-optimal parameter settings.

\begin{table}[t]
\centering
\caption{ACER / Balanced Accuracy values for a \gls{ctlstm} model with different
    color space utilization. Reported values apply for the validation subsets.}
\begin{tabular}{| l | c | c |} 
 \hline
 Utilized Color Spaces & SiW & OULU-NPU \\
 \hline\hline
 \ycbcr & 0.029 / 97.2\% & 0.1944 / 91.4\% \\
 \hline
 HSV AND \ycbcr & 0.013 / 98.8\% & 0.0897 / 95.5\% \\
 \hline
\end{tabular}
\label{table:lstm-color-space-utilization}
\end{table}

\subsection{Feature Impact}

We are interested in the impact that each feature has on the \gls{ctlstm} model.
For this purpose, we performed model trainings on the SiW dataset with only a single feature for
each training run.

We choose the following configuration for this experiment:
Dual-\gls{lstm} stack with 480 hidden units for the first \gls{lstm} layer and 240 hidden units for the second one;
10 frames per sample;
64 samples per batch;
30 epochs;
dropout frequencies: 0.2;
\gls{lbp} radius: $R=8$;
\gls{lbp} sample points: $P=8$;
color histogram buckets: 50;
color space: \ycbcr.

The results show that the color histogram feature, although performing moderately
well on its own with a balanced accuracy of 92.2\%, is easily surpassed by the
\gls{lbp} feature with a balanced accuracy of 99.4\%. Still, a combination of
both features achieves an even better performance of 99.6\%. We therefore
continue to use a combination of both features.

\subsection{Sample Length Impact}
Besides the utilized features, also the number of consecutive frames per sample
impacts the quantity of information that is available for classification. A
larger number of frames is assumed to result in better classification, but it
also slows down the authentication process and therefore negatively impacts the
user experience.

Our model is mainly tested with 16 frames that equals an authentication delay
of about half a second for cameras that record with 30 frames per second. In
order to verify the assumption stated above, we also conducted an experiment
with 30 frames per sample, i.e., an authentication delay of one second. This
model was trained for only 14 epochs but achieved considerable results on the test
subsets (see Table~\ref{table:lstm-sequence-length}). While these
results prove the above assumption correct, we do not define this model as the
optimal one because of the negative impact on user experience.

\begin{table}[t]
\centering
\caption{Performance of the \gls{ctlstm} model with
    optimal parameters and a sample length of 30 frames. Reported
    values apply for test subsets.}
\begin{tabular}{| l | c | c | c | c |} 
 \hline
 Dataset & Bal. Accuracy & ACER & APCER & BPCER \\
 \hline\hline
 SiW             & 99.04\% & 0.00798 & 0.0031178 & 0.01285 \\
 \hline
 OULU-NPU        & 95.52\% & 0.09037 & 0.1651 & 0.01564 \\
 \hline
 RECOD           & 99.94\% & - & - & 0.00055 \\
 \hline
\end{tabular}
\label{table:lstm-sequence-length}
\end{table}

\subsection{Resource Consumption}

With the parameters for both the features and the \gls{ctlstm} model set, it is
possible to determine the amount of data that must be transmitted to the
system. We assume that the model is located on a server that receives sample
data from mobile applications. By using 4 bytes per integer, 16 frames per
sample, 6 color channels, and 34 (\gls{lbp}) + 50 (color) histogram buckets,
the number of uncompressed bytes per sample is 8064  or 1KB per frame. With only histogram data
being used for inference, the sample size is independent of video resolution.
Unlike other approaches that rely on the transmission of complete video
streams, this method is remarkably lightweight.

\subsection{Dataset Impact}
\label{sec:eval:lstm:di}

We observe that high accuracy can be achieved for SiW and RECOD at relatively few epochs.
OULU-NPU, however, proves to be a more difficult
classification problem. Even for the human eye, it is difficult to
differentiate between certain live samples and attack recordings. For this reason, we
analyze the performance of the \gls{ctlstm} on this dataset in more detail by
recording the classification results for different dataset properties (such as
the attack type, the lighting condition, and the phone type).

We find that varying lighting conditions and attack types have only small impact
on the performance. However, the phone type is found to hold the highest variance with a
difference of over 15 percentage points between phone 2 (90.9\% balanced acc.)
and phone 4 (75.4\% balanced acc.). Yet, this cannot be explained with a difference
in the phone models, as both devices have cameras of similar properties (same
resolution and a nearly identical aperture size).

\subsection{Training with Optimal Parameters}

\begin{table}[t]
\centering
\caption{Performance of the \gls{ctlstm} model with optimal parameters. Reported values apply for test subsets.}
\begin{tabular}{| l | l | c | c | c |} 
 \hline
    Train Mode & Test Dataset & Bal. Accuracy & ROC AUC & ACER  \\
 \hline\hline
\multirow{3}*{Individual} & SiW             & 99.55\% & 99.95\% & 0.004  \\
\cline{2-5}
    & OULU-NPU        & 94.13\% & 98.01\% & 0.059 \\
\cline{2-5}
    & RECOD           & 98.54\% & 99.90\% & 0.015  \\
 \hline\hline
\multirow{3}*{Collective} & SiW             & 99.67\% & 99.96\% & 0.003  \\
\cline{2-5}
    & OULU-NPU        & 94.16\% & 97.92\% & 0.058 \\
\cline{2-5}
    & RECOD           & 97.53\% & 99.69\% & 0.025  \\
 \hline
\end{tabular}
\label{table:lstm-best-results}
\end{table}

With the combined findings of the sections above, we prepare a model with
the parameters that we know to work best. We choose the following configuration
for this experiment:
Dual-\gls{lstm} stack with 240 hidden units;
16 frames per sample;
32 samples per batch;
\gls{lbp} radius: $R=8$;
\gls{lbp} sample points: $P=32$;
color histogram buckets: 50;
color space: HSV and \ycbcr.

In the first run, \gls{ctlstm} is trained for each dataset individually. In a
second experiment, a single \gls{ctlstm} model is trained with all datasets.
The results of this experiment are listed in
Table~\ref{table:lstm-best-results}. With individual training, our model
achieves a
respective performance of 99.55\% and 98.54\% balanced
accuracy for the test subsets of SiW and RECOD.

There is no significant performance difference between
the individually and collectively trained models. This indicates that
\gls{ctlstm} is capable of classifying live samples and attacks from diverse
environments correctly.

\subsection{Cross-Dataset Evaluation}
\label{sec:cross-data}

In this experiment, a \gls{ctlstm} model is trained and tested with completely
different datasets such that the data used for testing was never seen during
training. The performance that is observed with this method is a strong
indicator for the model's capability to generalize across domains.

The \gls{ctlstm} model is trained with a combination of the datasets SiW, OULU-NPU,
and RECOD. The performance for
each dataset combination is reported in Table~\ref{table:lstm-cross-dataset}. The
best performance is achieved with \gls{ctlstm} models that are evaluated with SiW
and OULU-NPU data.

In addition to our proposed classification approach that uses only 16 frames
for classification, we also perform full-video classification for this
experiment (see right column in Table~\ref{table:lstm-cross-dataset}). This is
achieved by classifying 16-frame sequences individually and using the resulting
classifications for a majority vote on the whole video. Note that this mode of
classification resembles the approach of continuous authentication.
It can be observed that the model performance benefits from the capability to
view the whole video and vote on individual sequences.

\begin{table}[t]
\centering
\caption{HTER values of the \gls{ctlstm} model for cross-dataset testing.
    Training and testing is conducted on complete datasets. Per Video
    performance is achieved with majority voting.}
\begin{tabular}{| l | l | c | c |} 
    \hline
    Train & Test & 16 Frames & Per Video\\
    \hline
    \hline
    OULU-NPU \& RECOD & SiW & 0.157 & 0.083 \\
    \hline
    SiW \& RECOD & OULU-NPU & 0.274 & 0.265 \\
    \hline
    OULU-NPU \& SiW & RECOD & 0.359 & 0.322 \\
 \hline
\end{tabular}
\label{table:lstm-cross-dataset}
\end{table}

\subsection{Comparison to Related Work}

With this section we address the performance comparison of existing liveness
detection models with our approach. Multiple papers report their model
performance only on datasets that we do not have access to. Certain papers,
however, do report values for SiW and OULU-NPU protocols. We follow this approach
and train \gls{ctlstm} on SiW protocol 1 as well as OULU-NPU protocols 1 and 2.

In SiW protocol 1 only the first 60 frames of each video are seen during
training while the complete video is used for testing. The first 60 frames
mainly cover frontal view faces whereas
the rest of the video includes pose changes~\cite{liu2018learning}. In OULU-NPU
protocol 1, the model is trained with videos that were captured with two
specific lighting conditions. The test set consists exclusively of videos from
a third, unseen lighting condition. Similarly, protocol 2 uses two different
types of print and display attacks for training and testing. This simulates the
condition of unseen attacks.

The full comparison
for the datasets SiW and OULU-NPU is listed in
Table~\ref{table:related-work}. It must be noted that the expressiveness of
this comparison is
limited, because each approach uses a different amount of frames and it is
often unclear, if multiple classifications for the same video were merged to
one. To clarify, the performance reported for \gls{ctlstm} reflects the performance
video-wise as described in Section~\ref{sec:cross-data}.

It can be observed that \gls{ctlstm} experiences a performance drop in
comparison to training with the full dataset (see
Table~\ref{table:lstm-best-results}). Still, \gls{ctlstm} is able to halve the
error rate of the original color \gls{lbp} approach~\cite{boulkenafet2017oulu}.

\begin{table}[t]
\centering
\caption{Performance comparison of our model with related work. Reported values
    represent the ACER.}
\begin{tabular}{| l | c | c | c |} 
 \hline
    \multirow{2}*{Model} &
    SiW & \multicolumn{2}{c|}{OULU-NPU} \\
    \cline{2-4}
    & Protocol 1 & Protocol 1 & Protocol 2 \\
 \hline\hline
     Uncertainty Estimation~\cite{xu2021improving} 
     & 0.0003 & 0.002 & 0.011 \\
 \hline
     CDCN~\cite{cdcn} 
     & 0.0012 & 0.002 & 0.013 \\
 \hline
     STASN~\cite{yang2019face} 
     & 0.003 & 0.01 & 0.011  \\
 \hline
     \textbf{\gls{ctlstm}} 
     & 0.0039 & 0.097 & 0.076\\
 \hline
     RSGB \& STPM~\cite{fas_sgtd} 
     & 0.004 & 0.001 & 0.019  \\
 \hline
     CNN-RNN with \gls{rppg}~\cite{liu2018learning} 
     & 0.036 & 0.016 & 0.025 \\
 \hline
     ED-LBP~\cite{shu2021face} 
     & -- & 0.083 & 0.049 \\
 \hline
     DeepPixBiS~\cite{george2019deep} 
     & -- & 0.042 & 0.060 \\
 \hline
     Color \gls{lbp}~\cite{boulkenafet2017oulu} 
     & -- & 0.135 & 0.142  \\
 \hline
\end{tabular}
\label{table:related-work}
\end{table}

\section{Conclusion}
\label{sec:conclusion}

This work proposes a new approach to \acrlong{pad} that combines a previously
unused variant of the established \gls{lbp} feature with time-aware deep
learning based on \gls{lstm} layers. The design choice of the proposed
\gls{ctlstm} model maximizes the compatibility with available mobile devices
and minimizes the resource requirements.

All \gls{ctlstm} related parameters
are comprehensively described and analyzed. The model's performance is evaluated with
well-known \gls{pad} datasets. It is shown that the model categorizes attacks
and live recordings accurately at a sample size of only 16 frames. In
cross-dataset experiments it is observed that the model's generalizability can
be improved through majority voting of multiple samples, indicating that
continuous authentication may be advantageous in previously unseen settings.


\section*{Acknowledgment}



%\bibliographystyle{IEEEtranS}
\bibliographystyle{IEEEtranS} % We choose the "plain" reference style
\bibliography{main} % Entries are in the refs.bib file

\end{document}